\documentclass[sigconf]{acmart}

\usepackage{booktabs} % For formal tables

\usepackage{amsmath}
\usepackage{graphicx}
\usepackage{subcaption}
\usepackage{multirow}
\usepackage{enumitem}
\usepackage{natbib}
\usepackage{algorithm}
\usepackage{algorithmicx}
\usepackage{algpseudocode}

\newcommand{\argmax}[1]{\underset{#1}{\operatorname{arg}\operatorname{max}}\;}

\setcopyright{none}
\acmDOI{10.475/123_4}

\acmISBN{123-4567-24-567/08/06}

\acmConference[WWW'19]{ACM Woodstock conference}{May 2019}{San Francisco, California USA}
\acmYear{2019}
\copyrightyear{2019}

\acmArticle{4}
\acmPrice{15.00}

\begin{document}
\title{Layout Design for Intelligent Warehouse by Evolution with Fitness Approximation}

\author{Haifeng Zhang}
%\authornote{Peking University}
\affiliation{
  \institution{Peking University}
%  \streetaddress{P.O. Box 1212}
}
%\email{pkuzhf@pku.edu.cn}

\author{Zilong Guo}
\affiliation{
  \institution{Shanghai Jiao Tong University}
}
%\email{pkuzhf@pku.edu.cn}

\author{Han Cai}
\affiliation{
  \institution{Shanghai Jiao Tong University}
}

\author{Chris Wang}
\affiliation{
  \institution{Peking University}
}

\author{Weinan Zhang}
\affiliation{
  \institution{Shanghai Jiao Tong University}
}

\author{Yong Yu}
\affiliation{
  \institution{Shanghai Jiao Tong University}
}

\author{Wenxin Li}
\affiliation{
  \institution{Peking University}
}

\author{Jun Wang}
\affiliation{
  \institution{University College London}
}

%\keywords{ACM proceedings, \LaTeX, text tagging}

\renewcommand{\shortauthors}{Haifeng Zhang et al.}

\begin{abstract}
With the rapid growth of the express industry, intelligent warehouses that employ autonomous robots for carrying parcels have been widely used to handle the vast express volume. For such warehouses, the warehouse layout design plays a key role in improving the transportation efficiency. However, this  work is still done by human experts, which is expensive and leads to suboptimal results. In this paper, we aim to automate the warehouse layout designing process. We propose a two-layer evolutionary algorithm to efficiently explore the warehouse layout space, where an auxiliary objective fitness approximation model is introduced to predict the outcome of the designed warehouse layout and a two-layer population structure is proposed to incorporate the approximation model into the ordinary evolution framework. Empirical experiments show that our method can efficiently design effective warehouse layouts that outperform both heuristic-designed and vanilla evolution-designed warehouse layouts. 

%In this paper, we consider a robot warehouse optimization problem where the parcel-carry robots are controllable and the warehouse map is designable. We formulate the problem as a parametrised multi-agent environment that the environment and the agents share a common objective to optimize. We control agents by planning methods and focus on the design of the environment. We propose a evolutionary framework that searches the optimal environment parameter with the help of a continuously learning fitness approximation function. The ground-truth samples of the fitness function are generated by simulations of the agents acting in the environments. 
\end{abstract}

\maketitle

\section{Introduction}
% problem statement
% automatic warehouse design and challendges
The global express delivery industry has been a trillion market, serving the people's daily life around the world. In 2017, the industry revenue is 248 billion USD \citep{ibis2018global} and in China, particularly, the annual gross express volume has surpassed 30 billion USD since 2016 \citep{fan2017considerable}. During the recent two years, a new type of shipping warehouses, with intelligent robots sorting thousands of parcels per hour, emerged \citep{ChinaDaily2017robots}.
As shown in Figure~\ref{fig:real-warehouse} and \ref{fig:warehouse-env}, autonomous robots carry parcels across the warehouse and unload the parcels into the target holes which connect to the vehicles heading to the target destinations.
The layout of the warehouse, i.e. the matching of the holes and the target destinations, is usually designed by human experts. It can be challenging and also likely to be suboptimal, especially when the number of holes is large as shown in Figure~\ref{fig:warehouse-env}. Moreover, the demand of such warehouse layout design is not one-off, since the distribution of the parcel destinations is not fixed and the warehouse layout design should be adaptive to achieve the best performance.

%Inspired by the recent success achieved by machine learning techniques in replacing manual design such as neural network architectures \citep{zoph2016neural}, optimizer \citep{bello2017neural}, deep learning compiler stack \citep{chen2018tvm} and etc., we intend to automate the laborious warehouse layout design process. A big challenge in this task is that evaluating each designed warehouse layout would require to actually run agents in it, which is highly time-consuming even virtually. Thus, such a task requires the algorithm to be efficient in utilizing the expensive evaluations, which is lacking in many existing techniques \citep{real2017large,real2018regularized}. 

% propose our work
In this paper, we present an evolution-based method for automatically designing warehouse layout. To tackle the efficiency issue arising from time-consuming evaluation of each designed warehouse layout, we consider to train a neural network to predict outcomes of layouts without actually running agents in it, which is known as fitness approximation in the context of evolution \citep{jin2005comprehensive}. We further propose a novel two-layer population structure to incorporate the prediction model into the evolution framework for improving efficiency, which can be categorised as multiple-deme parallel genetic algorithms\citep{cantu1998survey}. Particularly, the higher layer consists of layouts that are actually evaluated and occupies a small fraction of the whole population while the lower layer contains layouts whose fitnesses are predicted by the learned model. Compared to existing methods for combining fitness approximation with evolution \citep{de2004hierarchical, hong2003acceleration}, the proposed two-layer evolutionary algorithm explicitly manages evaluated individuals and predicted individuals separately in two sub-populations and trains the approximation model online using the samples evaluated by the original fitness function. As such, the proposed method incorporates fitness function approximation into the multiple-deme parallel genetic algorithm naturally. Moreover, within an evaluation of a designed warehouse layout, we can observe not only the final outcome but also additional agent trajectories that comprise hidden information about the causes of the outcome. To take advantage of such additional information to improve the quality of the prediction model, we construct an auxiliary objective, i.e. to predict the heatmap of the environment where each individual value is the total number of visits of a point. 

% results
Our experiments of designing warehouse layouts demonstrate improved efficiency and better performance compared to both manual design and vanilla evolution-based methods without fitness approximation. Such a two-layer evolution-based environment optimization framework is promising to be applied onto various environment design tasks.

\section{Related Work}
% environment design
There are many real-world scenarios that can be regarded as environment design problems, ranging from game-level design with a desired level of difficulty \citep{togelius2011search}, shopping space design for impulsing customer purchase and long stay \citep{penn2005complexity} to traffic signal control for improving transportation efficiency \citep{ceylan2004traffic}. In a recent work, \citep{zhang2018learning} formulates these environment design problems using a reinforcement learning framework. In this paper, we focus on a new environment design scenario, i.e. warehouse layout design, emerging from the rapidly growing express industry. 

Traditional warehouse design problems can be categorised to three levels, strategic level, tactical level and operational level \citep{rouwenhorst2000warehouse}. At the strategic level, long-term decisions are considered, including the size of a warehouse \citep{roll1989determining} and the selection of component systems \citep{oser1996design, keserla1994analysis}. At the tactical level, medium term decisions are made, such as the layout of a conventional warehouse \citep{bassan1980internal, berry1968elements}. At the operational level, detailed control policies are studied, e.g. batching \citep{elsayed1983computerized} and storage policies \citep{goetschalckx1991optimal}. The problem discussed in this paper is about warehouse layout design, which is at the tactical level traditionally. However, in the era of big data, the layout of warehouse could be adaptive to the changes of the external environment. Specifically, the layout of the warehouse could be redesigned at intervals according to the changing destination distribution of the parcels. Thus, this problem is better to be categorised as a operational level problem.

For solving this problem, we adopt evolutionary algorithms. As getting a guiding signal means evaluating the designed objective in the target task, which would result in unacceptable computational resource requirement for scenarios where evaluation is expensive. To reduce the amount of expensive evaluations on real data needed before a satisfying result can be obtained, some works propose to learn a model to predict the outcome of a designed objective without actually running on real data \citep{baker2018accelerating,liu2017progressive}. Similar idea has been explored in the field of evolution and is known as fitness approximation \citep{jin2005comprehensive}. Due to the inaccuracy of fitness approximation, it is essential to use the approximation model together with the original fitness function \citep{grierson1993optimal, ratle1998accelerating}. To incorporate the fitness model into the simulation-based evolutionary algorithms, individual-based \citep{bull1999model} and generation-based \citep{ratle1998accelerating} methods are studied. Differently, our approach explicitly manages two sub-populations whose individuals are evaluated by the approximation model and the original fitness function respectively. Similar approaches are known as multiple-deme parallel genetic algorithms \citep{cantu1998survey}. Our work can be classified as a multiple-deme parallel genetic algorithm with a two-layer sub-population topology to balance exploitation and exploration. 

\section{Problem Definition} \label{sec:problem}
In this section, we formulate the environment design problem and introduce the particular robotic warehouse environment. We fix the agent policy in the robotic warehouse environment and focus on the remaining task, assigning destinations to the holes, which can be viewed as an environment design problem.

\subsection{Environment Design} \label{sec:env-design}

In many scenarios, there are $n$ agents taking actions in a designable environment, such as cars running in a transportation system, consumers shopping in a mall, and so on. Denote the $i^{th}$ agent's policy as $\pi_i$ and the environment is parametrized as $\mathcal{M}_\theta=\langle S, A, T_\theta, R_\theta, \lambda \rangle$, where $S,A,T_\theta,R_\theta,\lambda$ denote state space, action space, transition function, reward function and reward discount respectively. After the agents play in the environment in an episode, a joint trajectory $H= \langle s_1, a_1, s_2, a_2, ... \rangle$ is produced and a cumulative reward $G_i$ is given to the $i^{th}$ agent, where $s_t$ and $a_t$ denote state and joint action respectively. Moreover, the objective of the environment designer is given as $O(H)$, whose function form can be defined specifically, and the designer intends to design an optimal environment to maximize the expectation of its objective
\begin{align}
\theta^* = \argmax{\theta} \mathbb{E}[O(H)|\mathcal{M}_\theta; \pi_{1\ldots n}].
\label{eq:original-general-problem}
\end{align}
Note that the randomness of $H$ is derived from the possible randomness of $\pi_i$ when selecting actions. 

\begin{figure}[t]
	\centering
	\begin{subfigure}{0.35\linewidth}
		\centering
        \includegraphics[height=.98\columnwidth]{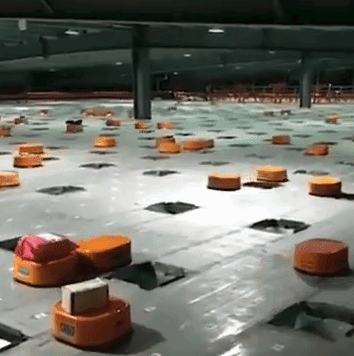}
        \caption{}
        \label{fig:real-warehouse}
	\end{subfigure}
	\hfill
	\begin{subfigure}{0.6\linewidth}
		\centering
        \includegraphics[height=0.6\columnwidth]{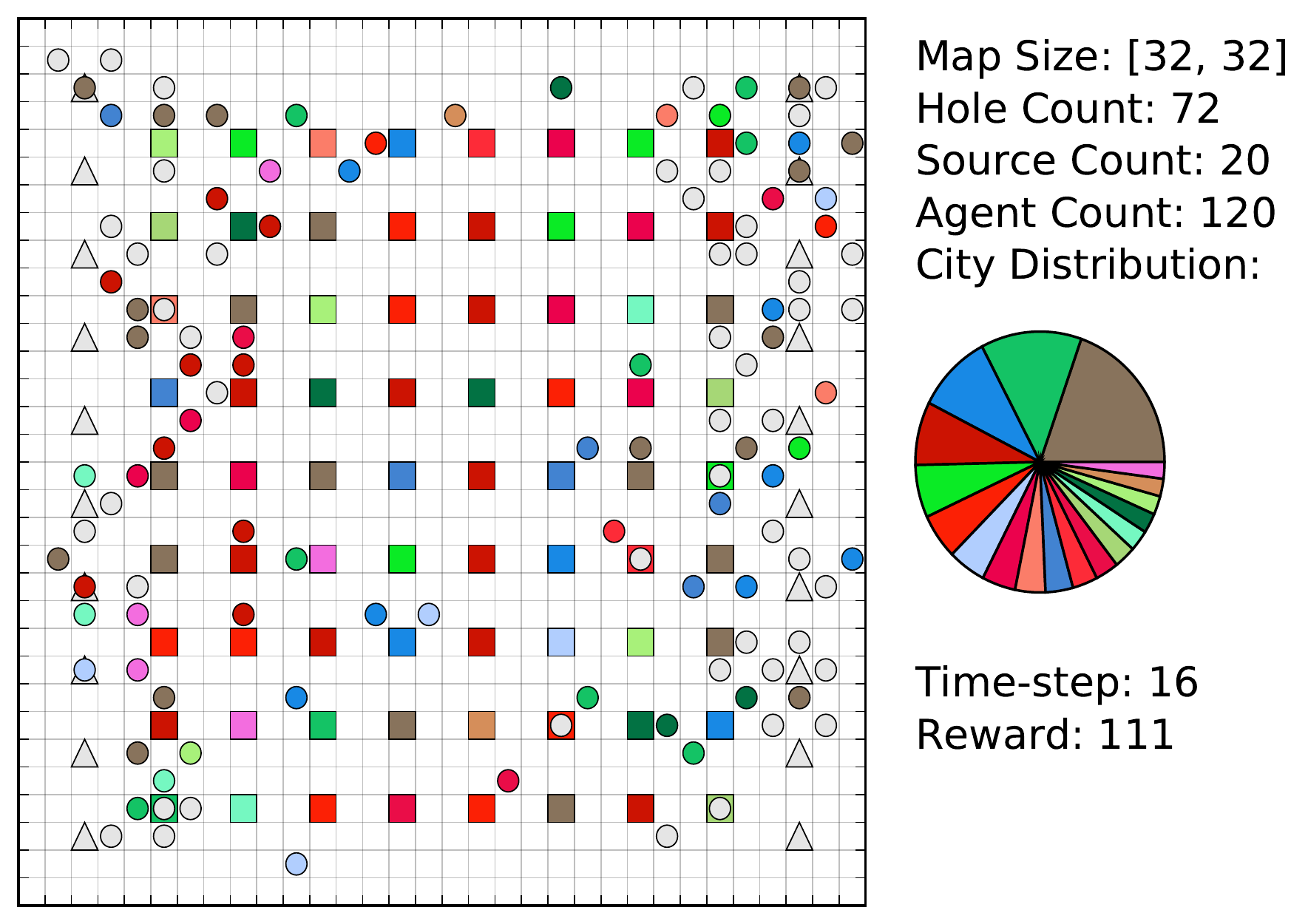}
        \caption{}
        \label{fig:warehouse-env}
	\end{subfigure}	
	\caption{(a) Real-world robotic warehouse for parcel sorting (screenshot from \citep{ChinaDaily2017robots}). (b) Robotic warehouse environment. The triangles stand for the sources where parcels emerge. The circles stand for the robots carrying the parcels. The squares stand for the holes for the agents to put into the parcels. The squares are colored according to which destination the parcels coming into will go to. The agents repeatedly take a parcel with a color (destination) from a source to a hole with the same color. The objective is to maximize the total number of the parcels processed by the agents in a fixed period.}
\end{figure}

\subsection{Robotic Warehouse Environment}

In this paper, we consider a robotic warehouse environment abstracted from a real-world express system as shown in Figure~\ref{fig:real-warehouse}, where there is a warehouse for sorting parcels from a mixed input stream to separate output streams according to their respective destinations. The sorting process is done by the robots carrying parcels from the input positions (sources) to the appropriate output positions (holes) in the plane warehouse as Figure~\ref{fig:warehouse-env} illustrates. In order to maximize the efficiency of sorting, we should set the robots' cooperative pathfinding algorithm and assign the destinations to the holes. In this task, the agents share a common reward $G$ and the environment also takes $G$ as its design objective, i.e. $O(H) = G$. We set $\pi_\phi$ as a joint policy model for the agents. As such, the problem is formulated as
\begin{align}
\theta^*, \phi^* = \argmax{\theta, \phi} \mathbb{E}[G|\mathcal{M}_\theta, \pi_{\phi}].
\label{eq:general-problem}
\end{align}
For solving Eq.~(\ref{eq:general-problem}), we should firstly set a sound cooperative pathfinding algorithm $\pi_{\phi^*}$ for the robots. After, we focus on optimizing the environment parameter $\theta$, i.e. optimizing the layout of the warehouse (the assignment of the destinations to the holes) via
\begin{align}
\theta^* = \argmax{\theta} \mathbb{E}[G|\mathcal{M}_\theta; \pi_{\phi^*}].
\end{align}
Note that the demand of such environment layout design is not one-off. Since the external variables (such as the destination distribution of the parcels) may be changing, the best layout of the warehouse is changing accordingly. Thus, the layout of the warehouse should be redesigned at intervals, which gives a reason to find an efficient layout design approach.

\subsection{Detailed Environment Description} \label{sec:experiment-desc}

The warehouse is abstracted as a grid containing $h \times w$ cells. Among them, $n_s$ cells are sources and $n_h$ cells are holes, whose locations $l_{s}^{1..n_s}, l_{h}^{1..n_h}$ are given. There are $n_r$ robots available to carrying parcels from sources to holes. Each cell is only for one robot to stand.

In each time-step, each robot is able to take a move to an adjacent cell. When an empty robot moves into a source, it loads a new parcel whose destination follows a distribution over $n_d$ destinations (cities) with the proportions $p_1,p_2,...,p_{n_d}$. On the other hand, when a loaded robot moves into a hole with the destination that is as the same as the loading parcel's, it unloads the parcel into that hole.  That is to say, the rates of input and output flows are not restricted in our setting. Parcels are always sufficient when a robot moves into a source.

Our objective is to sort as many parcels as possible in a given time period $T$. We could achieve this objective by designing the layout of the warehouse, i.e. assigning the proper destinations to the holes. Specifically, we should determine the parameter $\theta=\langle \theta_1,\theta_2,...,\theta_{n_h} \rangle$ of the environment $\mathcal{M}_\theta$, where $\theta_i \in \{1..n_d\}$ for $i=1..n_h$. Intuitively, the assignment of the destinations to the holes will affect the robots' paths and hence the efficiency of the whole warehouse. 

The notations defined in this section are listed in Table~\ref{table:notation}.

\subsection{Problem Complexity}

For the problem defined above, the scale of the layout assignment space is $n_d^{n_h}$, where $n_h$ denotes the number of the holes and $n_d$ denotes the number of the parcel destinations. Since the robot pathfinding algorithm works like a black box to evaluate each layout assignment, it is hard to determine a global optimum without exploring the solution space completely. Thus, this optimization problem is an exponential time problem. Even for a small setting, such as $n_h=20, n_d=5$, the number of the assignments is as large as about $100$ trillion, which is hard to be explored completely.

\begin{table}
\centering
\caption{Notations and descriptions}
\label{table:notation}
\resizebox{0.8\columnwidth}{!}{
\begin{tabular}{c|l|c}
\hline
Notation & Description & Type\\
\hline
$h$ & Height of warehouse & Input \\
$w$ & Width of warehouse & Input \\
$n_s$ & Number of source cells & Input \\
$n_h$ & Number of hole cells & Input \\
$l_s^{1..n_s}$ & Locations of source cells & Input \\ 
$l_h^{1..n_h}$ & Locations of hole cells & Input \\
$n_r$ & Number of robots & Input \\
$n_d$ & Number of parcel destinations & Input \\
$p_{1..n_d}$ & Proportions of parcel destinations & Input \\
$T$ & Length of timestep & Input \\
$\theta_{1..n_h}$ & Assignment of destinations to holes & Output \\
\hline
\end{tabular}
}
\vspace{-10pt}
\end{table}

\subsection{Robot Pathfinding Algorithms} \label{sec:problem-ware-design}

\begin{figure}[t]
	\centering
	\begin{subfigure}{0.32\linewidth}
		\centering
        \includegraphics[height=0.98\columnwidth]{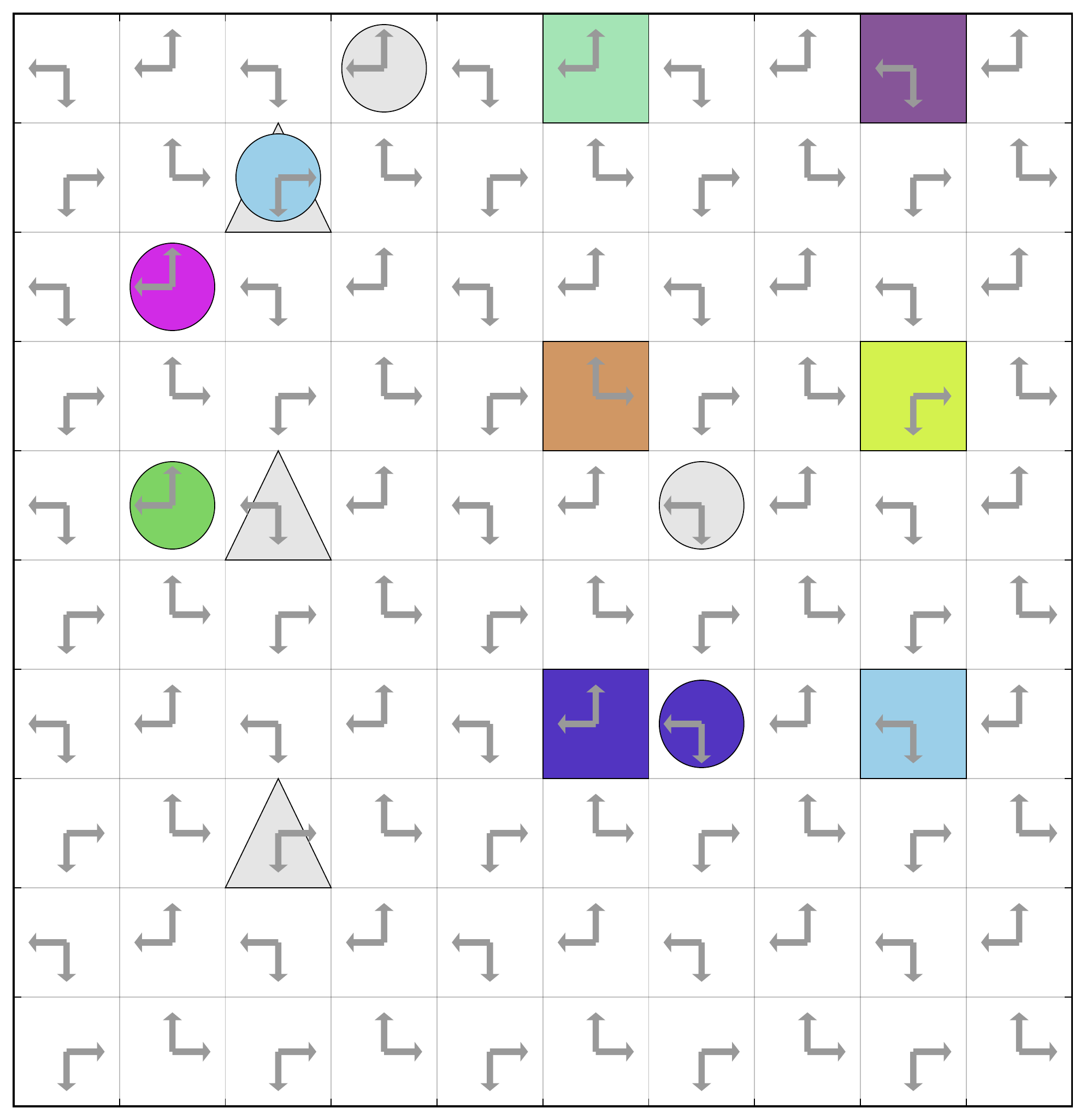}
        \caption{}
        \label{fig:road}
	\end{subfigure}
	\hfill
	\begin{subfigure}{0.32\linewidth}
		\centering
        \includegraphics[height=0.98\columnwidth]{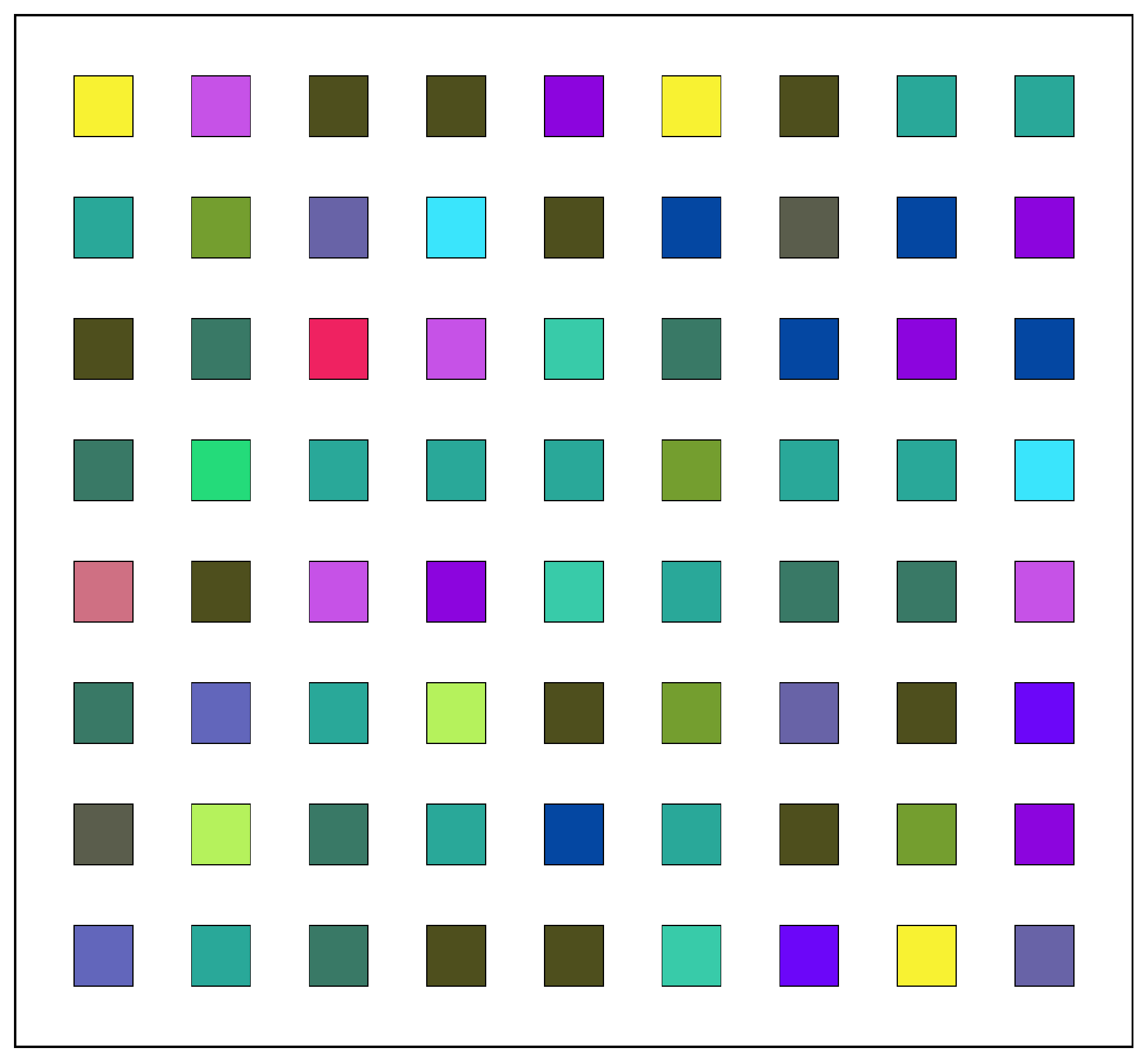}
        \caption{}
        \label{fig:ea-sample}
	\end{subfigure}
	\hfill
	\begin{subfigure}{0.32\linewidth}
		\centering
        \includegraphics[height=0.98\columnwidth]{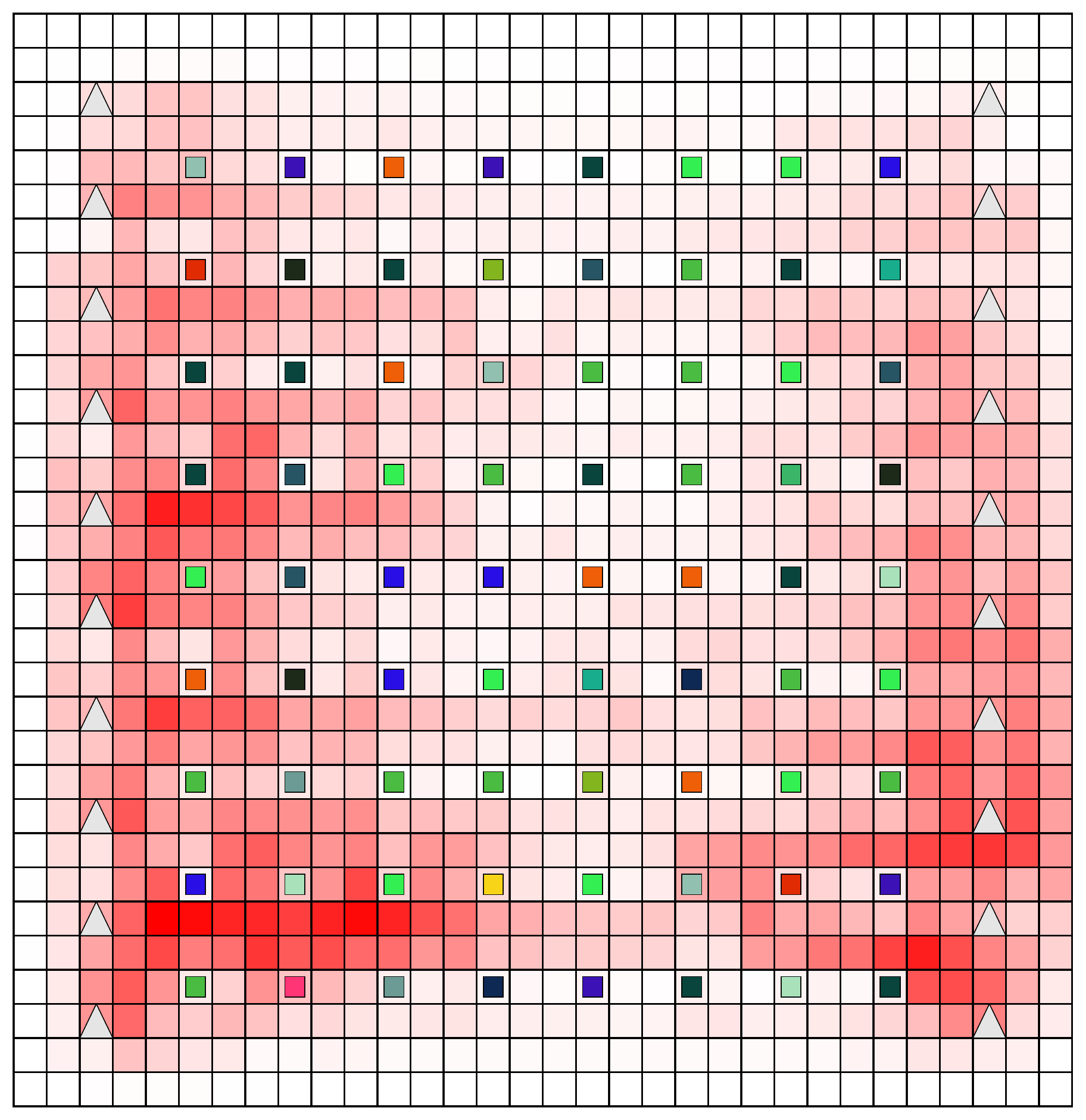}
        \caption{}
        \label{fig:heatmap}
	\end{subfigure}	
	\caption{(a) An illustration of one-way roads: i) the odd-row cells allow moving right and forbid moving left, while the even-row cells allow moving left and forbid moving right; ii) the odd-column cells allow moving down and forbid moving up, while the even-column cells allow moving up and forbid moving down. The left-down cell is in Row 1 and Column 1. (b) A layout sample as an individual in the evolutionary algorithm. (c) An example of the heatmap.}
\vspace{-10pt}
\end{figure}

In our problem, the robot pathfinding algorithm is fixed. As the robots are quite dense in the real-world warehouse, jam prevention is the key point. We considered two cooperative pathfinding algorithms with jam prevention design. The first one adopts WHCA* \citep{silver2005cooperative} as a planner, which searches the shortest path from an origin to a destination for each robot in turn and ensures non-collision. The second algorithm is a greedy one, which guides the robots by a look-up table in each position and reduces conflicts by setting one-way roads in the map as illustrated in Figure~\ref{fig:road}. We studied these two algorithms and the results showed that the greedy one has a significant advantage on time complexity and a minor disadvantage on performance. Due to the large simulation demand for testing environment parameter, we selected the time-saving greedy algorithm as the agent policy in our experiments. However, the proposed warehouse layout design solution can work with other robot pathfinding algorithm as well.

\section{Solution} \label{sec:solution}

In this section, we first introduce an evolution framework for automatically designing warehouse layout, and then present the auxiliary objective fitness approximation and the two-layer population structure for improving the efficiency.

\subsection{Evolution with Robot Policy Simulation} \label{sec:solution-baseline}

Under the evolution framework, we maintain a population containing $n$ warehouse layout individuals, i.e. assignments of the destinations to the holes (Figure~\ref{fig:ea-sample}), and evolve the population for $n_g$ generations. Within each generation, we perform crossover, mutation and selection in order: 

\begin{itemize}
\item In the \textbf{crossover} phase, we randomly select $c$ pairs of samples. For each pair of samples, we splice their holes from two matrices to two lines respectively. Then, we randomly select a common breakpoint for both lines and cross the two lines just like chromosomal crossover. Finally we generate two square matrices by reshaping the two lines.
\item In the \textbf{mutation} phase, we randomly select $m_1$ samples generated in the crossover phase. For each sample, we randomly select $m_2$ holes and randomly permute their destinations. 
\item In the \textbf{selection} phase, we evaluate the generated samples in the crossover and mutation phases by robot policy simulations, then merge the original and the generated samples. The best $n$ ones are selected for the next generation. 
\end{itemize}

\subsection{Two-layer Evolutionary Algorithm with Fitness Approximation} \label{sec:solution-proposed}

In this section, we propose a novel evolutionary algorithm that trains an auxiliary objective fitness function to evaluate a large population for providing promising individuals to a small population evaluated by simulations.

\subsubsection{Auxiliary Objective Fitness Approximation}

\begin{figure}
	\centering
	\includegraphics[width=1\linewidth]{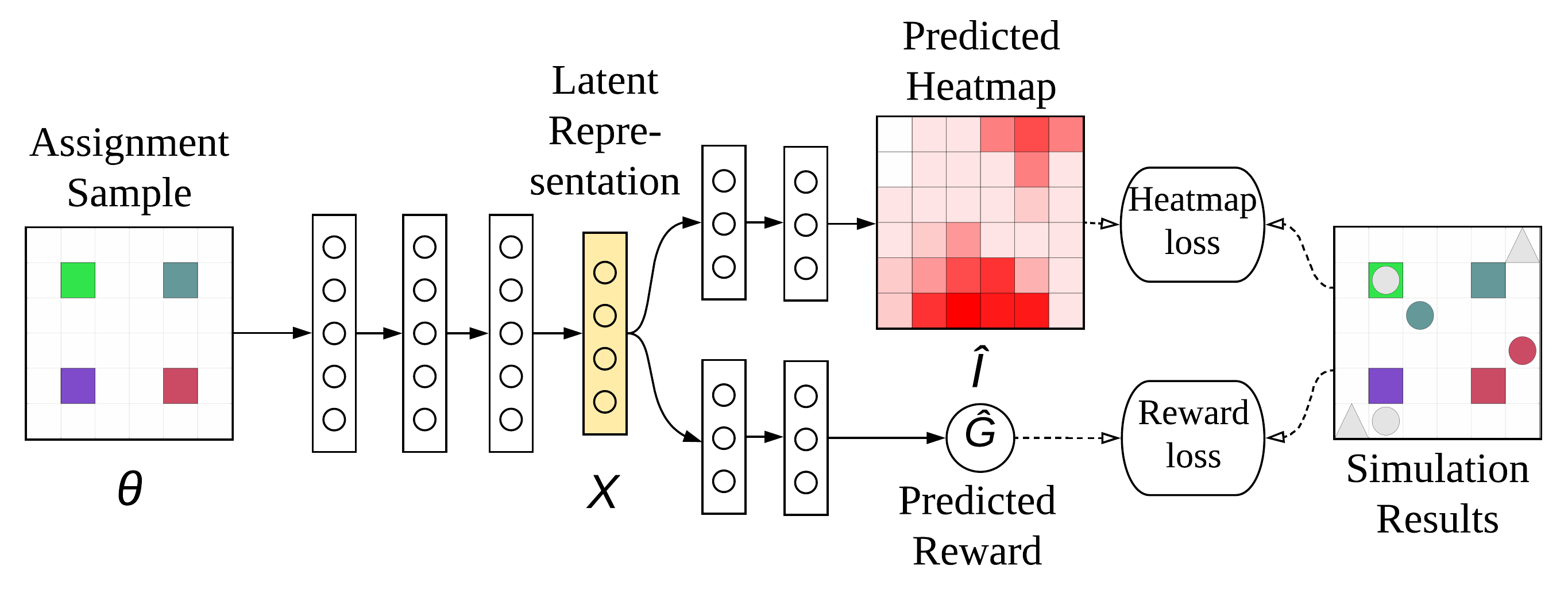}
	\caption{An illustration of the process of evaluating an assignment sample $\theta$. First, the latent representation $X$ is learned via shared deep layers. Then based on $X$, separated layers are built to predict heatmap $\hat{I}$ and reward $\hat{G}$ respectively. Two loss functions are calculated based on the difference between the prediction and the simulated results.}
	\label{fig:aux-obj-network}
\end{figure}

In practise, the simulation of robots performing in the environment is time-consuming. A promising way to reducing the simulation time is to use an approximation function to compute fitness:
\begin{align}
f_G(\theta) = \hat{G} \approx \mathbb{E}[G|\mathcal{M}_\theta; \pi_{\phi^*}],
\end{align}
where $f_G$ is the fitness approximation function, $\theta$ is a sample of environment parameter and $\hat{G}$ is the predicted fitness of $\theta$, whose learning target is the expectation of the reward $G$.

Moreover, since a simulation generates a trajectory $H$ in addition to the reward $G$, we consider utilizing $H$ to help training fitness function $f_G$. Although $G$ is the exact objective for fitness function to learn, we may extract additional information $I(H)$ from $H$ that helps training the fitness function, under the assumption that $G$ and $I$ are correlated. We set an auxiliary training objective and use a neural network to capture this:
\begin{align}
f(\theta) &= \langle f_I(f_X(\theta)), f_G(f_X(\theta)) \rangle = \langle \hat{I}, \hat{G} \rangle \\ \nonumber
&\approx \langle \mathbb{E}[I(H)|\mathcal{M}_\theta, \pi_{\phi^*}], \mathbb{E}[G|\mathcal{M}_\theta, \pi_{\phi^*}] \rangle,
\end{align}
where $f$ is a neural network consisting of three sub-networks: $f_X$ is the bottom network that captures the common features and outputs $X$; $f_I$ and $f_G$ are the two separate networks on the top of $X$ that predict $\hat{I}$ and $\hat{G}$ respectively. 

In the robotic warehouse layout design problem, $\theta$ represents the assignment of the destinations to the holes and $H$ represents the movements of the robots. Furthermore, we define $I$ as the heatmap of the movements as Figure~\ref{fig:heatmap} shows. Intuitively, the distribution of busy areas should be correlated with the efficiency of sorting and the reward. The process of learning the fitness function in the warehouse layout problem is illustrated in Figure~\ref{fig:aux-obj-network}. 

Since obtaining simulation samples is time-consuming, we train the fitness model online. Specifically, the fitness model is trained with the samples simulated along the process of the evolutionary algorithm. There is no pre-training in our approach.

\subsubsection{Two-layer Population}

\begin{figure}
	\centering
	\includegraphics[width=1\linewidth]{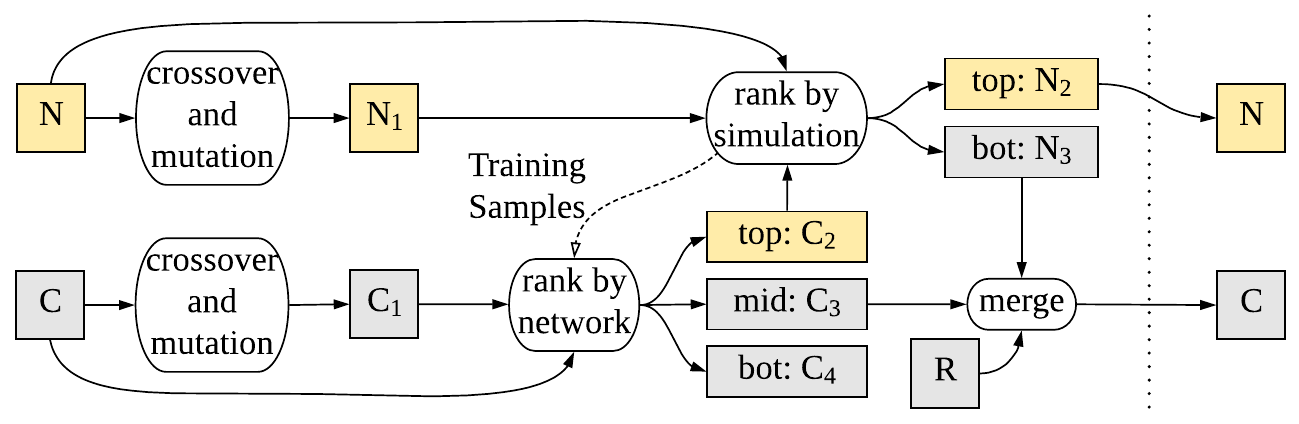}
	\caption{The process of the two-layer population evolutionary algorithm in a single generation. The yellow and grey squares stand for the populations who have been (or will be) evaluated by simulation and fitness model respectively.}
	\label{fig:two-layer-ea}
\end{figure}

The fitness model provides a less accurate but more speedy evaluation than the simulation. These property indicates that the simulation is better to find the local optimum exactly and the fitness model is better to explore the global space speedily. For the standard simulation-based evolution, mutation rate is usually set small enough to ensure convergence within an acceptable time, thus the search space is relatively local. Therefore, we consider incorporating the fitness model into the standard simulation-based evolution as an additional part for exploring the global space. 

Specifically, we maintain two sub-populations. The first one is of the same size as the population set in the standard simulation-based evolution. Also, the individuals in the first sub-population are evaluated by simulations. The second sub-population is multiple times larger than the first one and the samples in it are evaluated by the fitness model. We view the second sub-population as a candidate population whose top individuals have a chance of joining the first sub-population. On the other hand, the bottom individuals in the first sub-population may be moved to the second sub-population. We name the first-layer sub-population noble and the second civilian. Noble population and civilian population evolve separately while keeping a channel for migration.  

\begin{algorithm}[t]
\caption{Two-layer Evolutionary Algorithm with Fitness Approximation (a literal expalnation of Figure~\ref{fig:two-layer-ea})}\label{algo:tlea}
\begin{algorithmic}[1]
\Require noble population $N$, civilian population $C$, untrained fitness model $f$, empty simulation sample set $S$
\For {each generation}
\State generate $N_1$ from $N$ by crossover and mutation;
\State generate $C_1$ from $C$ by crossover and mutation;
\State rank $C \cup C_1$ by $f$ to generate top population $C_2$, middle population $C_3$ and bottom population $C_4$;
\State evaluate $N_1$ and $C_2$ by simulation and add the results to $S$;
\State rank $N \cup N_1 \cup C_2$ by the simulation score to generate top population $N_2$ and bottom population $N_3$;
\State generate random population $R$ and discard $C_4$;
\State pass $N_2$ to the next generation as $N$;
\State pass $N_3 \cup C_3 \cup R$ to the next generation as $C$;
\State update $f$ using $S$. 
\EndFor
\end{algorithmic}
\end{algorithm}

In detail, the two-layer population evolves as Figure~\ref{fig:two-layer-ea} and Algorithm~\ref{algo:tlea} show. In general, $N$ and $C$ maintain individuals evaluated by the simulation and the fitness model respectively. In each generation, migration takes place. Specifically, $C_2$ from the civilian layer go up to the noble layer and $N_3$ from the noble layer go down to the civilian layer. In addition, the civilian layer discards the worst population $C_4$ and absorbs randomly generated population $R$. 

There are $9$ parameters related to the proposed two-layer evolutionary algorithm. They are noble population number $|N|$, civilian population number $|C|$, crossover rate $c_N, c_C$, mutation rate $m_N, m_C$, $|C_2|$ for the number of civilian individuals migrate to the noble layer, $|R|$ for the number of the randomly generated individuals, and $n_u$ for the number of model updates in each generation. Other variables can be determined by these parameters. In each generation, $|N_1|+|C_2|$ simulations, $n_u$ model updates and $|C_1|$ model predictions are performed. Since the time cost of training the network and use it to predict is negligible compared to the simulations (see Table~\ref{table:time-cost}), the time complexity of the two-layer evolutionary algorithm for $n_g$ generations is $O(n_g(|N_1|+|C_2|))$.

\section{Experiment} \label{sec:experiment}

We set up a virtual intelligent warehouse environment based on real-world settings and test our proposed approach comparing to the baselines. Our experiment is repeatable and the source code is provided in the supplementary.

\subsection{Experiment Settings}

\emph{Environment.} We test our proposed approach in $20 \times 20$ maps. The positions of the sources and holes are set as the real-world scenarios. The detailed parameters are given in Table~\ref{table:warehouse-parameter}. The destination distributions are set according to long-tail functions to reflect reality. 
%$T$ in $32 \times 32$ map is smaller than in $20 \times 20$ map because simulation for larger map costs more time. 
In our experiments, the reward is defined as the sum of parcel loading times and unloading times (roughly two times as the number of parcels processed).

%\begin{table}[]
%\centering
%\caption{Environment parameter settings.}
%\label{table:warehouse-parameter}
%\small
%\begin{tabular}{|c|c|c|c|c|c|c|c|}
%\hline
%No. & $h$ & $w$ & $n_s$ & $n_h$ & $n_r$ & $n_d$ & $T$ \\ \hline
%1 & 20 & 20 & 12 & 20 & 60 & 5 & 1000 \\ \hline
%2 & 32 & 32 & 20 & 72 & 80 & 18 & 300 \\ \hline
% & \multicolumn{7}{c|}{Destination Distribution} \\ \hline
%1 & \multicolumn{7}{c|}{$M(0.367, 0.267, 0.2, 0.133, 0.033)$} \\ \hline
%2 & \multicolumn{7}{c|}{$M(p_{1\ldots 18}), p_i = \frac{1}{(2+i) \times \sum_i \frac{1}{2+i}}, i=1\dots 18$} \\ \hline
%\end{tabular}
%\vspace{-10pt}
%\end{table}

\begin{table}[]
\centering
\caption{Environment parameter settings.}
\label{table:warehouse-parameter}
\small
\begin{tabular}{|c|c|c|c|c|c|c|c|}
\hline
$h$ & $w$ & $n_s$ & $n_h$ & $n_r$ & $n_d$ & $T$ & $p_{1..n_d}$\\ \hline
20 & 20 & 12 & 20 & 60 & 5 & 1000 & $0.367, 0.267, 0.2, 0.133, 0.033$\\ \hline
\end{tabular}
\vspace{-10pt}
\end{table}

\emph{Robots.} As introduced, we adopt a greedy algorithm as the cooperative pathfiding algorithm for the robots. Firstly, we set one-way roads in the map as Figure~\ref{fig:road} shows to avoid opposite-directional conflicts, while right-angled conflicts are avoided by setting priority. On the one-way roads, the robots decide moves by a look-up table containing $h \times w \times (n_s + n_h)$ records, each of which indicates the first step towards a particular source or hole from a particular cell. 

\emph{Baselines.} We test $5$ baselines to compare with our proposed two-layer evolutionary algorithm (\textbf{TLEA}). \textbf{Random}: The holes are assigned with random destinations uniformly. \textbf{Heuristic}: Destinations select holes in turns according to their proportions. For example, if $10\%$ parcels are going to destination A, then A select $10\%$ of the holes. This process start from the destination with the most proportion. Each destination greedily selects each hole that minimizes the sum of the average distance from the sources to the selected holes. \textbf{Simu}: The evolutionary algorithm with simulations as introduced in the Solution section. \textbf{SimuInd}: An implementation of the individual-based evolution control algorithm \citep{bull1999model}. This approach maintains a single large population for evolution whose individuals are evaluated by the fitness model. In each generation, the best individuals evaluated by the fitness model are evaluated by the simulation once again. The fitness model is trained online with the samples produced by the simulations. \textbf{SimuGen}: An implementation of the generation-based evolution control algorithm \citep{ratle1998accelerating}. This approach also maintains a single large population as SimuInd. The difference is that SimuGen uses the simulations intensively in a generation and uses the fitness model in the next several generations.

\emph{Hyper-parameters.} To ensure fairness, for Simu, SimuInd, SimuGen and TLEA, the number of generation is set as $60$ and the number of simulations in each generation is set as $200$. The model update and prediction times are also fixed as $5000$ and $10000$ respectively for SimuInd, SimuGen and TLEA. The population of Simu is $100$; in each generation $200$ individuals are generated by crossover; $50$ of them are mutated. For SimuInd and SimuGen, the populations are $5000$; $10000$ are generated by crossover in each generation; $2500$ of them are mutated. For the TLEA, $|N|,|C|,c_N,c_C,m_N,m_C,|C_2|,|R|,n_u$ are set to be $100, 5000, 1, 1, 0.25, 0.25, 50, 2500, 5000$ respectively.

\emph{Fitness model.} Our network is composed of three sub-networks $f_X$, $f_I$, $f_G$. The output of $f_X$ is used for the input of $f_I$ and  $f_G$. $f_X$ has two fully connected layers whose output is a vector that can be reshaped to match the size of map. Then, a 2D transposed convolution layer follows.$f_I$ has one transposed convolution layer to generate the heat map. And $f_G$ contains three fully connected layers to predict the reward. All the layers except the output layers have a ReLU activation function. The loss functions for the two outputs are set to be MSE. The first two fully connected layers have 128, 400 units respectively. The first 2D transposed convolution layer have 16 filters. And the second one has one filter. The three fully connected layers for reward prediction have 256, 128 and 1 unit respectively.

\emph{Hardware.} We use two computers with an Intel core i7-4790k and an Intel core i7-6900k respectively. The one with 4790k also has an extra Nvidia Titan X GPU.

%\begin{table}
%\centering
%\vspace{-0pt}
%\caption{Performance of Random, Heuristic, Simu, SimuInd, SimuGen and TLEA.}
%\label{table:exp-result}
%\small
%\begin{tabular}{|l|c|c|}
%\hline
% & $20 \times 20$ & $32 \times 32$ \\ \hline
%Random & 4757(-11.7\%) & 1545(-13.5\%) \\ \hline
%Heuristic & 5386 & 1786 \\ \hline
%Simu & 5572(+3.5\%) & 1899(+6.3\%) \\ \hline
%SimuInd & 5605(+4.1\%) & 1925(+7.8\%) \\ \hline
%SimuGen & 5499(+2.1\%) & 1790(+0.2\%) \\ \hline
%TLEA  & \textbf{5646}(+4.8\%) & \textbf{1941}(+8.7\%) \\ \hline
%\end{tabular}
%\end{table}

\begin{table}
\centering
\vspace{-0pt}
\caption{Performance of Random, Heuristic, Simu, SimuInd, SimuGen and TLEA. The algorithms are repeatedly performed for $10$ runs. The reward samples pass the Shapiro-Wilk test to be normal. T-tests are performed for TLEA against Simu, SimuInd and SimuGen. The statistical results show that the superiority of TLEA is significant. }
\label{table:exp-result}
\footnotesize
\begin{tabular}{|l|c|c|c|c|c|c|}
\hline
& Random & Heuristic & Simu & SimuInd & SimuGen & TLEA \\ \hline
Reward & 4757 & 5386 & 5572 & 5605 & 5499 & \textbf{5646} \\ \hline
T Score & - & - & 5.8778 & 2.7708 & 5.8782 & - \\ \hline
P-Value & - & - & $7 \times 10^{-6}$ & $6.3 \times 10^{-4}$ & $7 \times 10^{-6}$ & - \\ \hline
\end{tabular}
\vspace{-10pt}
\end{table}

\begin{figure}
	\centering
	\begin{subfigure}{0.32\linewidth}
		\centering
        \includegraphics[height=1\columnwidth]{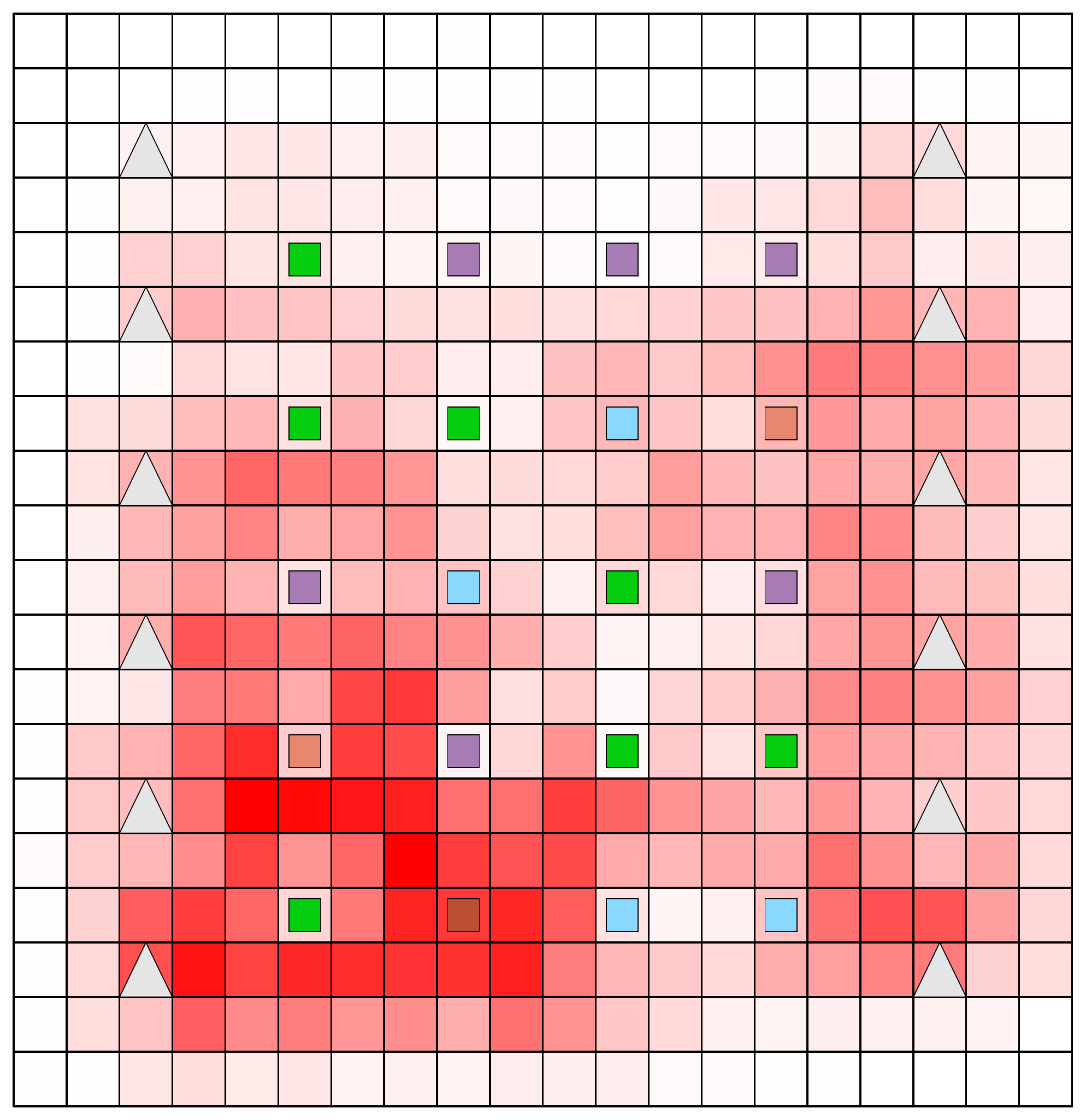}
        \caption{Random}
        \label{fig:result_random_20}
	\end{subfigure}
	\begin{subfigure}{0.32\linewidth}
		\centering
        \includegraphics[height=1\columnwidth]{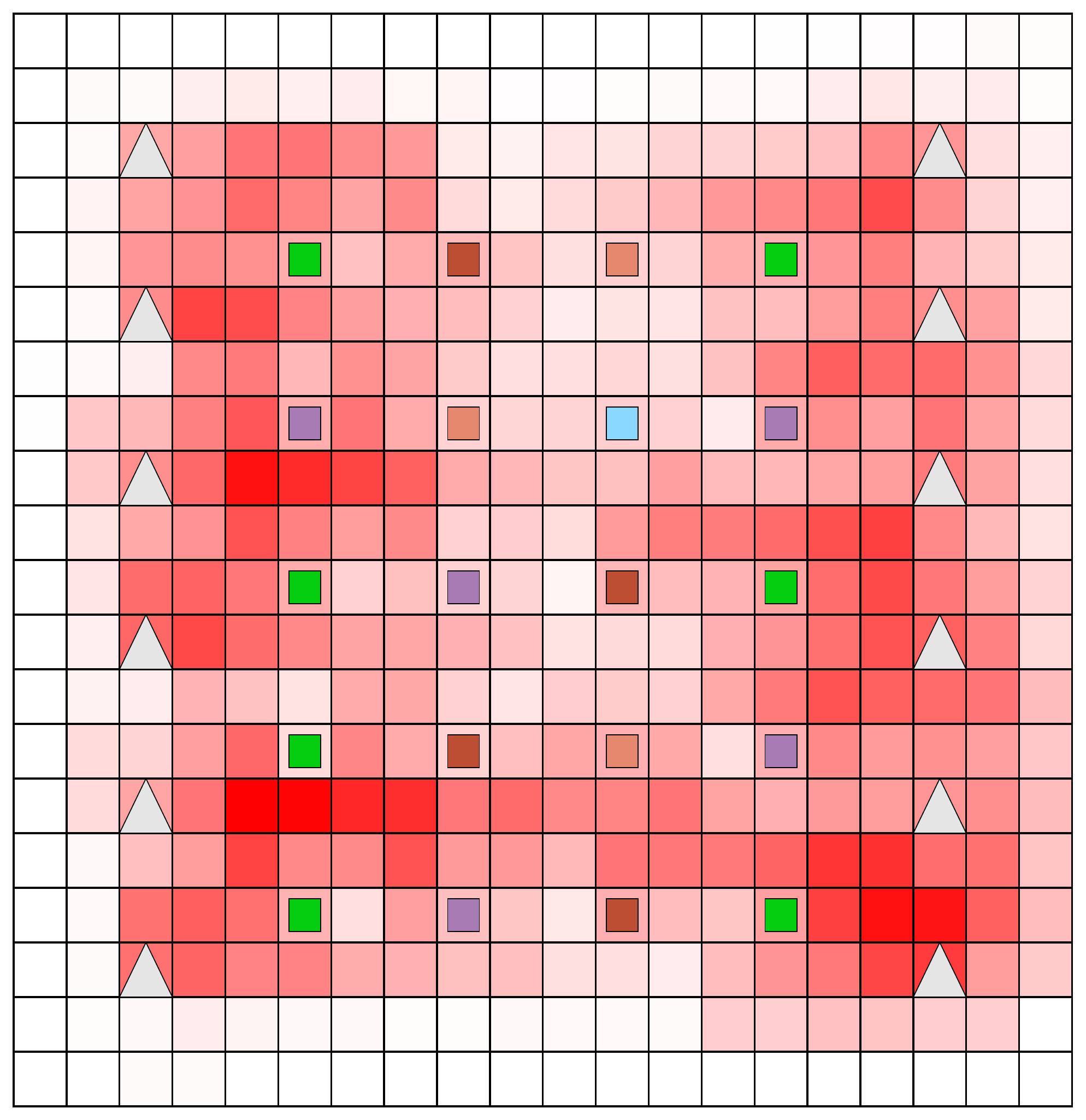}
        \caption{Heuristic}
        \label{fig:result_heuristic_20}
	\end{subfigure}
	\begin{subfigure}{0.32\linewidth}
		\centering
        \includegraphics[height=1\columnwidth]{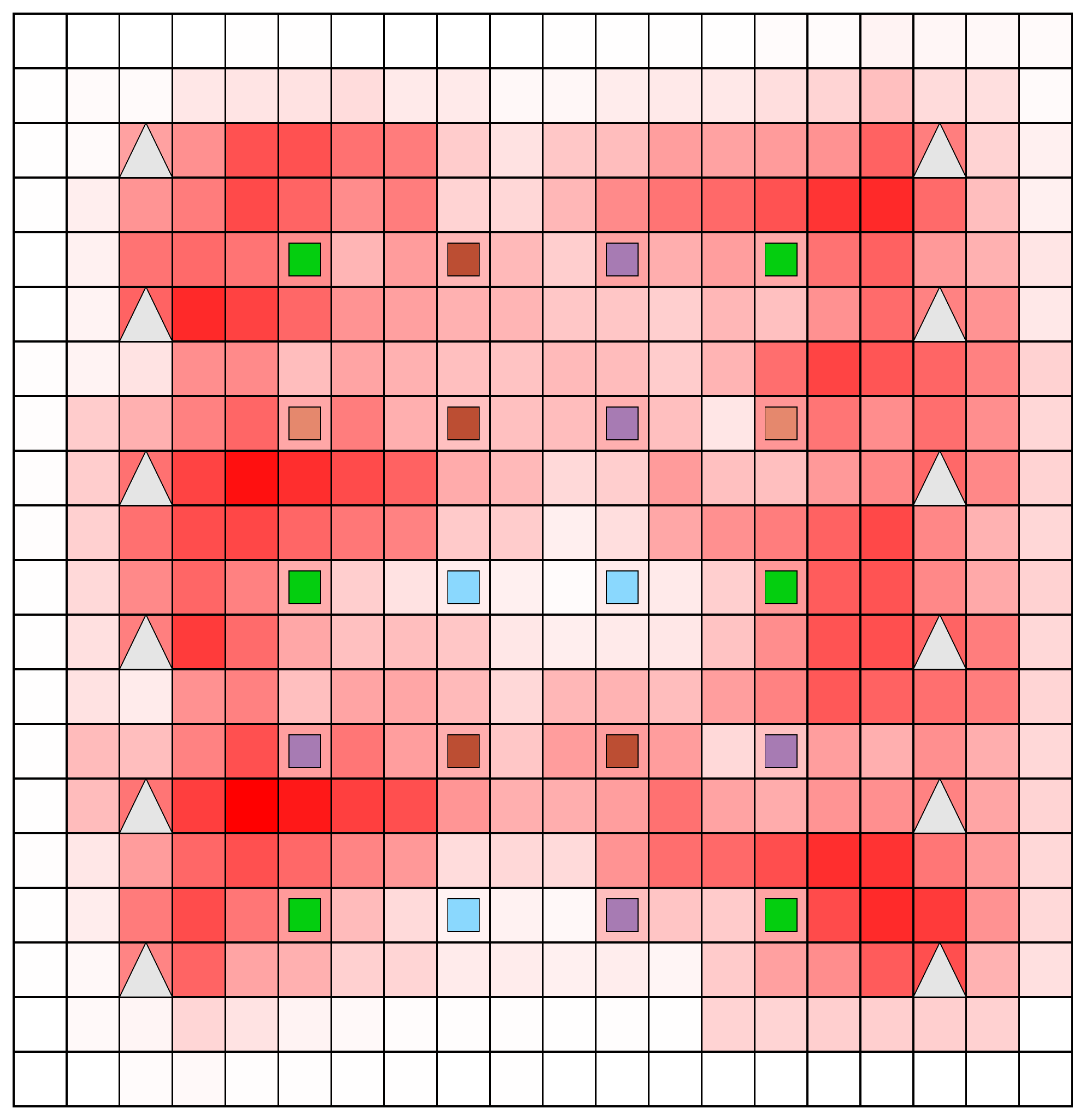}
        \caption{Simu}
        \label{fig:result_simu_20}
	\end{subfigure}
	\begin{subfigure}{0.32\linewidth}
		\centering
        \includegraphics[height=1\columnwidth]{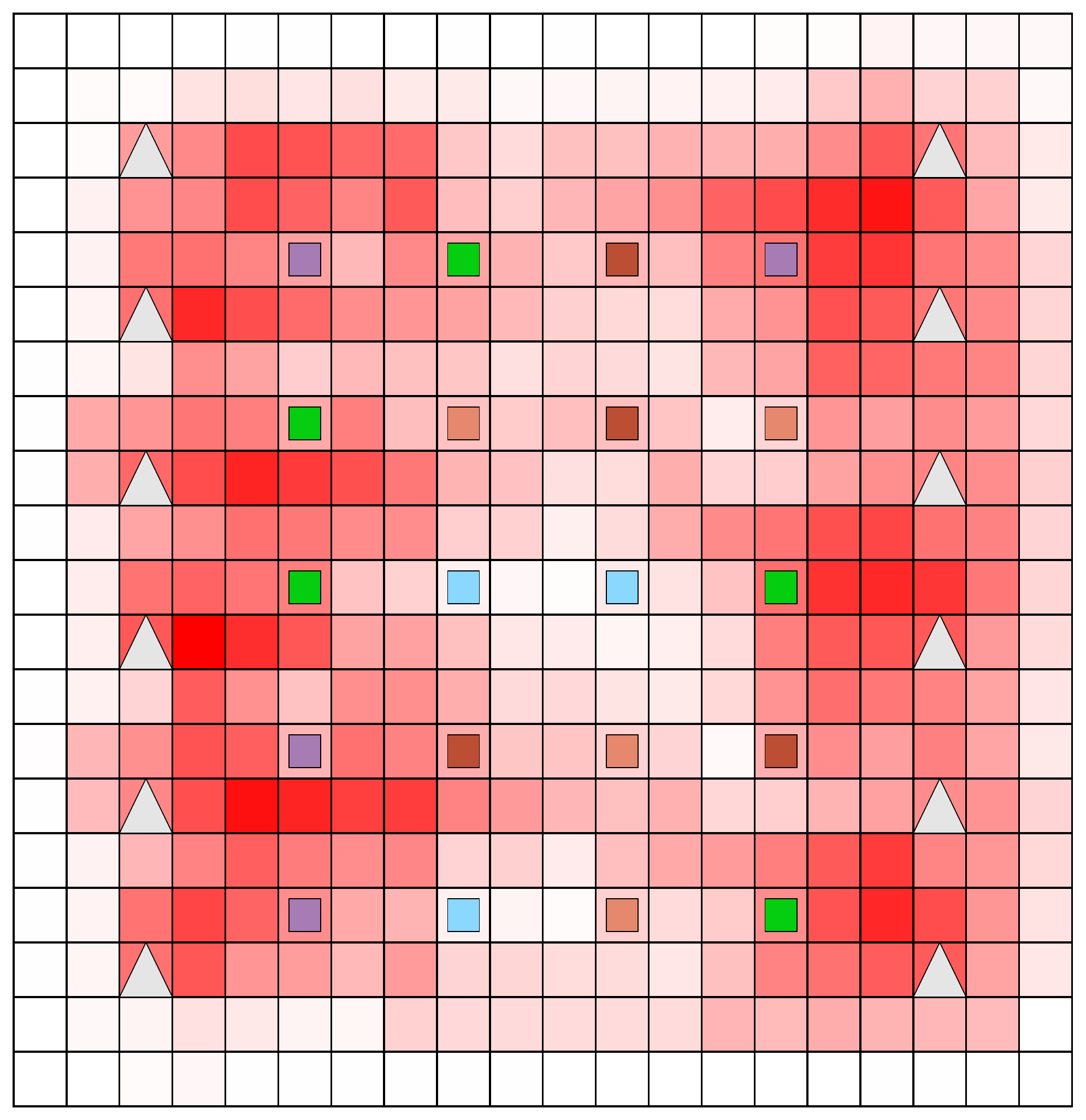}
        \caption{SimuInd}
        \label{fig:result_simuind_20}
	\end{subfigure}
	\begin{subfigure}{0.32\linewidth}
		\centering
        \includegraphics[height=1\columnwidth]{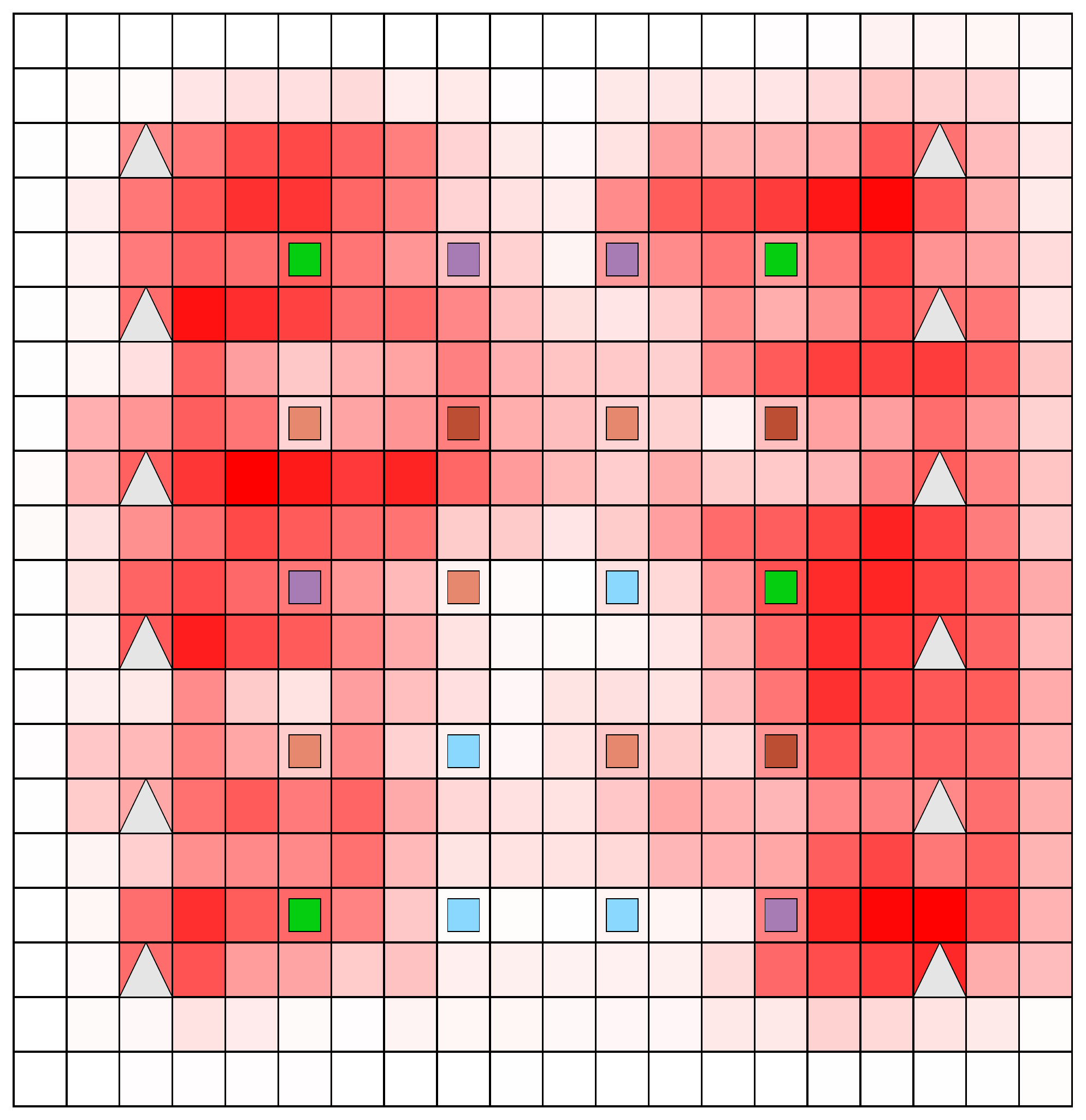}
        \caption{SimuGen}
        \label{fig:result_simugen_20}
	\end{subfigure}
	\begin{subfigure}{0.32\linewidth}
		\centering
        \includegraphics[height=1\columnwidth]{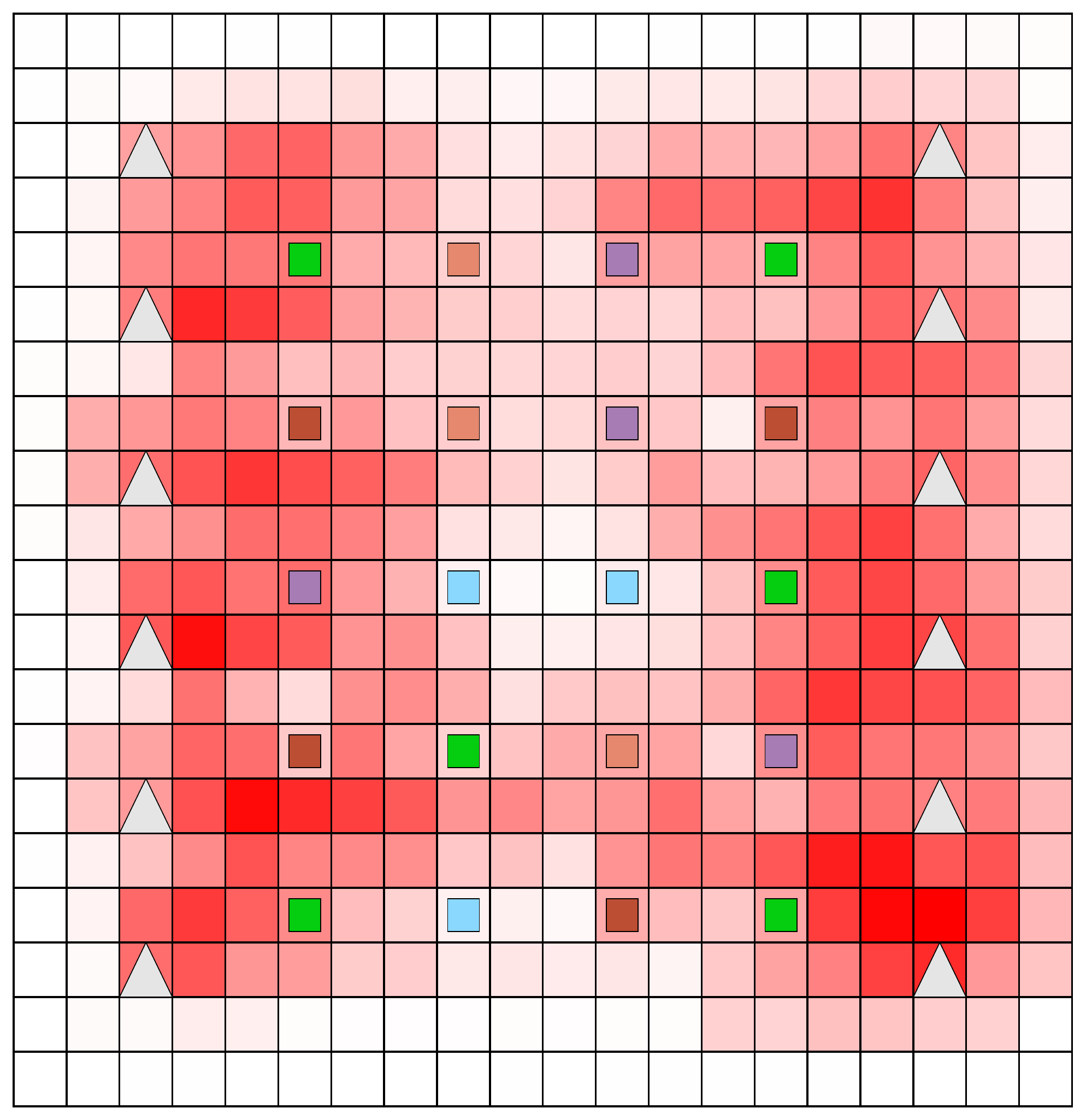}
        \caption{TLEA}
        \label{fig:result_tlea_20}
	\end{subfigure}
%	\vspace{-5pt}
	\caption{Environments designed by Random, Heuristic, Simu and TLEA.}%\vspace{-10pt}
	\label{fig:result}
	\vspace{-10pt}
\end{figure}

\subsection{Results}

We perform the baselines and TLEA. The results are shown in Table \ref{table:exp-result}. We find Heuristic is fairly high compared to Random but is inferior to evolutionary algorithms. Moreover, TLEA outperforms all the baselines. 

Figure~\ref{fig:result} shows the layouts designed by the baselines and TLEA with the heatmaps. We can see that the tracks of the robots running in the maps of TLEA are better balanced, indicating that there are less traffic jams. 

Figure~\ref{fig:converge} shows the learning curves. Since SimuInd and SimuGen mix the individuals evaluated by the simulation and the fitness model, their current best individuals may be the over-estimated ones by the inaccurate fitness model, which may lead to discarding the real best individuals. TLEA solves this problem by separating the two populations and ensure that the real best individual is always kept in the noble population. 

In addition, TLEA and Simu are more stable than SimuInd and SimuGen, because the temporary best individual may be evaluated by the fitness model in SimuInd and SimuGen, which may be corrected by the simulation in later generations. The slight fluctuations of Simu and TLEA are caused by the variance of the simulations, which results in that the best samples can be over-estimated (which is much slighter than the fitness model) and would be averaged by extra simulations in later generations.

%\begin{table}[t]
%\centering
%\caption{Comparison of WHCA* and Greedy Algorithm}
%\label{table:pathfinding}
%\small
%\begin{tabular}{|c|c|c|c|c|c|c|}
%\hline
%\begin{tabular}[c]{@{}c@{}}Agent\\ Count\end{tabular} & 
%\begin{tabular}[c]{@{}c@{}}WHCA*\\ Reward\end{tabular} & 
%\begin{tabular}[c]{@{}c@{}}Greedy\\ Reward\end{tabular} & 
%Difference & 
%\begin{tabular}[c]{@{}c@{}}WHCA*\\ Time\end{tabular} & 
%\begin{tabular}[c]{@{}c@{}}Greedy\\ Time\end{tabular} & 
%Difference \\ \hline
%40 & 3455.4 & 3112.2 & -9.93\% & 35.088 & 2.208 & -93.71\% \\ \hline
%80 & 5958.4 & 5332.8 & -10.50\% & 58.752 & 5.212 & -91.13\% \\ \hline
%120 & 7395 & 6729.2 & -9.00\% & 93.258 & 7.522 & -91.93\% \\ \hline
%\end{tabular}
%\end{table}

\subsection{Discussions} \label{sec:exp-analysis}

%\emph{Cooperative pathfinding algorithm selection.} We test WHCA* and Greedy in $32 \times 32$ map with agent number ranging from $40$ to $120$. The performance is given in Table~\ref{table:pathfinding}. As it shows, Greedy is $10$ times faster  and $10\%$ worse in performance than WHCA*. Although we select Greedy for the experiment, the proposed algorithm can also work with WHCA* or other pathfinding algorithms.

\emph{Time cost.} The time costs of the tested algorithms are listed in Table~\ref{table:time-cost}. It shows that the time cost proportion of the fitness model is less than $5\%$. In out experiment, we just ignore the time difference between Simu and other algorithms.

\begin{table}[]
\centering
\caption{Time cost comparison. The average time costs for simulation, model update and model predicting are $2.62$s, $2.42$ms and $1.06$ms respectively. The number of generations is $60$ for all the algorithms.}
\label{table:time-cost}
\small
\begin{tabular}{|l|c|c|c|c|}
\hline
 & Simulation & Model Update & Model Predicting & Time \\ \hline
Simu & $12k$ & $0$ & $0$ & $8.73h$\\ \hline
SimuInd & $12k$ & $300k$ & $600k$ & $9.11h$\\ \hline
SimuGen & $12k$ & $300k$ & $600k$ & $9.11h$\\ \hline
TLEA & $12k$ & $300k$ & $600k$ & $9.11h$\\ \hline
\end{tabular}
\vspace{-10pt}
\end{table}

\emph{Effectiveness of heatmap.} We evaluate randomly generated samples by the simulations and use them to train the fitness functions with and without heatmaps as auxiliary objective. We compare MSE and Pearson Correlation of them in Table~\ref{table:exp-heatmap}, which shows that heatmap provides significant improvement to the fitness function. 

\begin{table}[]
\centering
\caption{Comparison of fitness functions with and without heatmap.}
\label{table:exp-heatmap}
\small
\begin{tabular}{|c|c|c|c|c|}
\hline
\multirow{2}{*} {\begin{tabular}[c]{@{}c@{}}Sample\\ Number\end{tabular}} & \multicolumn{2}{c|}{MSE} & \multicolumn{2}{c|}{\begin{tabular}[c]{@{}c@{}}Pearson Correlation\end{tabular}} \\ \cline{2-5} 
 & w/o & Heatmap & w/o & Heatmap \\ \hline
5000 & 4.36 & 2.89(-33.72\%) & 0.277 & 0.519(+87.36\%) \\ \hline
10000 & 2.75 & 1.59(-42.18\%) & 0.405 & 0.687(+69.63\%) \\ \hline
20000 & 1.67 & 0.69(-58.68\%) & 0.766 & 0.908(+18.54\%) \\ \hline
\end{tabular}
\end{table}

\emph{Simulation allocation.} Since simulations are scarce resources when running evolutionary algorithm, the allocation of simulations between the noble layer and the civilian layer is important. Moreover, it also determines the migration rate between the two layers. We test different $\frac{|N_1|}{|N_1|+|C_2|}$, the ratio of simulations allocated to the noble layer, and find that $0.75$ is a proper setting (see Table~\ref{table:simulation-proportion}), which means three fourths simulations are allocated to ensure the accuracy of the noble layer and one fourth simulations are allocated to give chances to the civilian layer.

\begin{table}[]
\centering
\caption{Simulation allocation analysis.}
\label{table:simulation-proportion}
\small
\begin{tabular}{|c|c|c|c|c|}
\hline
Noble Proportion & 0.25 & 0.5 & 0.75 & 1 \\ \hline
Reward & 5629 & 5634 & \textbf{5646} & 5581 \\ \hline
\end{tabular}
\end{table}

\begin{figure}[t]
	\centering
	\begin{subfigure}{0.49\linewidth}
		\centering
        \includegraphics[height=0.7\columnwidth]{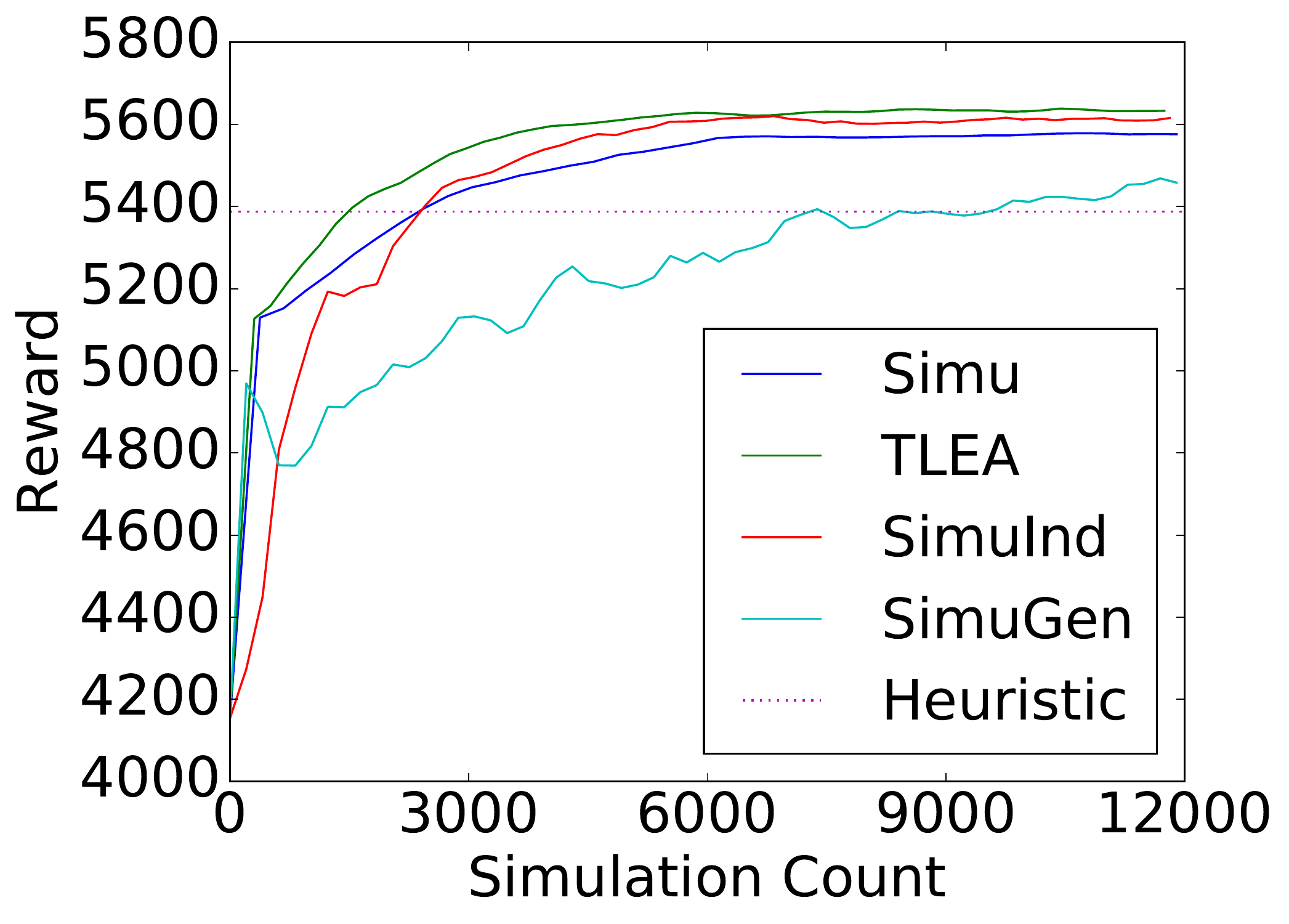}
        \caption{}
        \label{fig:converge}
	\end{subfigure}
	\hfill
	\begin{subfigure}{0.49\linewidth}
		\centering
        \includegraphics[height=0.7\columnwidth]{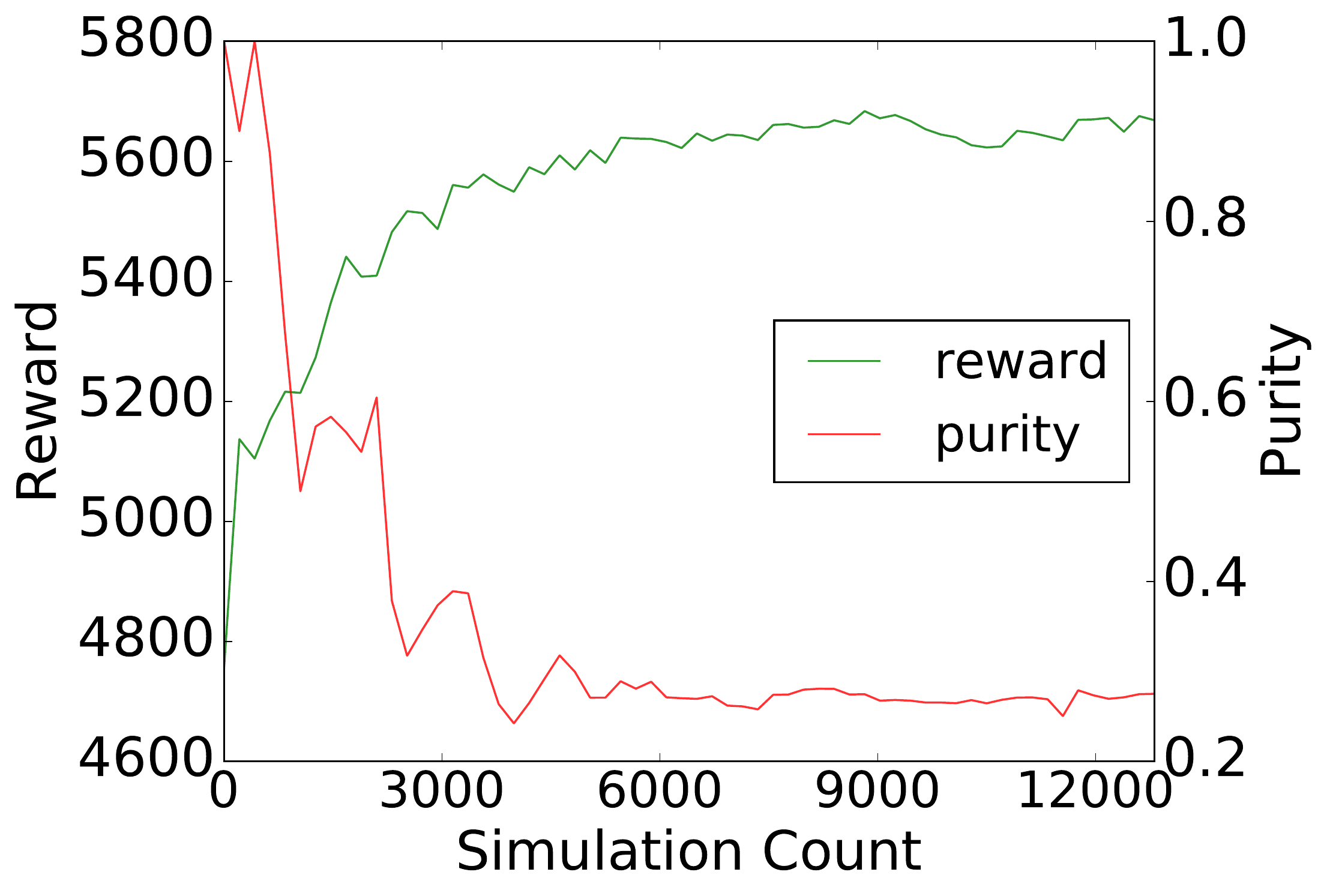}
        \caption{}
        \label{fig:purity}
	\end{subfigure}	
	\caption{(a) Learning curves averaged over $10$ runs. The Y-axis is the reward received by the best individual in each population. (b) Impact of civilian population for a particular run. Initially, the purity of each individual in the noble population is set to be $1$ and each civilian is set to be $0$. During the evolution, each child's purity is the mean of its parents' purity.}
\end{figure}

\emph{Impact of civilian population.} We are interested in how much contribution has the civilian population made to the evolution of the noble population. We calculate a number named purity that measures how much the evolved noble population inherits from the initial noble population. As Figure~\ref{fig:purity} shows, the purity of the noble population declines rapidly along with the increasing of the reward (fitness). Finally, civilian population contributes more than $70$ percent to the noble population.

\section{Conclusion}

In this paper, we study the problem of automatic warehouse layout design. The proposed two-layer evolutionary algorithm takes advantage of a fitness approximation model, augmented with an auxiliary objective of predicting the heatmap. Our approach enhances the exploration of the evolutionary algorithm with the help of the fitness model. The experiments demonstrates the superiority of our approach over the heuristic and the traditional evolution-based methods. For future work, we would apply the proposed two-layer evolutionary algorithm to other environment design scenarios, such as shopping mall design, game design and traffic light control.
%In this paper, we study the robot warehouse design problem and propose the two-layer evolutionary algorithm. We incorporate fitness approximation function into the evolution framework for evaluating a larger population evolving along with the original generation-evaluated population. Moreover, we train the fitness function using the heatmap of the robots' movement as an auxiliary objective. The experiments demonstrate the effectiveness of the proposed algorithm.

%Our two-layer evolutionary algorithm can be applied to other environment design scenarios, such as shopping mall design, game-level design and traffic light control. In the future, we intend to study the marriage of evolutionary algorithm and reinforcement learning for the environment design problem. 


%\input{samplebody-conf}

\bibliographystyle{ACM-Reference-Format}
\bibliography{storage-env}

\end{document}